\documentclass[10pt,twocolumn,letterpaper]{article}

\usepackage{iccv}
\usepackage{times}
\usepackage{epsfig}
\usepackage{graphicx}
\usepackage{amsmath}
\usepackage{amssymb}

\usepackage{multirow}

\usepackage{subfig}

\usepackage[breaklinks=true,bookmarks=false]{hyperref}

\iccvfinalcopy 

\def\iccvPaperID{9451} 

\ificcvfinal\fi

\begin{document}

\title{Shift-and-Balance Attention}

\author{Chunjie Luo, Jianfeng Zhan, Tianshu Hao, Lei Wang, Wanling Gao\\
Institute of Computing Technology, Chinese Academy of Sciences\\
Beijing, China\\
{\tt\small luochunjie,zhanjianfeng,haotianshu,wanglei\_2011,gaowanling@ict.ac.cn}
}

\maketitle
\ificcvfinal\thispagestyle{empty}\fi

\begin{abstract}
Attention is an effective mechanism to improve the deep model capability. Squeeze-and-Excite (SE) introduces a light-weight attention branch to enhance the network's representational power. The attention branch is gated using the Sigmoid function and multiplied by the feature map's trunk branch. It is too sensitive to coordinate and balance the trunk and attention branches' contributions. To control the attention branch's influence, we propose a new attention method, called Shift-and-Balance (SB). Different from Squeeze-and-Excite, the attention branch is regulated by the learned control factor to control the balance, then added into the feature map's trunk branch. Experiments show that Shift-and-Balance attention significantly improves the accuracy compared to Squeeze-and-Excite when applied in more layers, increasing more size and capacity of a network. Moreover, Shift-and-Balance attention achieves better or close accuracy compared to the state-of-art Dynamic Convolution.

\end{abstract}

\section{Introduction}

Deep neural networks (DNNs) have received great successes in many areas of machine intelligence. Modern state-of-art networks \cite{Simonyan2015VeryDC} \cite{He2016DeepRL} \cite{Huang2017DenselyCC} \cite{Szegedy2016RethinkingTI} \cite{Zoph2018LearningTA} require high computational and memory resources to improve accuracy. That increases the costs of large-scale applications. The  resource requirements also limit the usages of DNNs on mobile and embedded devices. As a result, there has been rising interest in designing efficient architectures of neural networks \cite{Iandola2017SqueezeNetAA} \cite{Howard2017MobileNetsEC} \cite{Sandler2018MobileNetV2IR} \cite{Howard2019SearchingFM} \cite{Tan2018MnasNetPN} \cite{Zhang2018ShuffleNetAE} \cite{Ma2018ShuffleNetVP} \cite{Zhang2017InterleavedGC} \cite{Xie2018IGCV2IS} \cite{Sun2018IGCV3IL}. 

Attention is an effective mechanism to enhance the model capability with a little computational cost.
Squeeze-and-Excite (SE) \cite{Hu2018SqueezeandExcitationN} introduces a light-weight attention branch that enhances the network's representational power by modeling channel-wise relationships. The attention branch is gated using Sigmoid function and multiplied by the feature map's trunk branch. The trunk branch is scaled by the coefficient generated by the attention branch.
Although the gate function Sigmoid constrains the attention branch to a certain degree, scaled attention is too sensitive to coordinate and balance the two branches' contributions. 
In extreme situations, it makes the whole channel inactivated when Sigmoid is saturated on the side of zero. That is a waste of channels, especially for light-weight networks where the channels are few. Sigmoid function also hinders the gradient backpropagation of the truck branch at training time.
Conditionally Parameterized Convolution (CondConv) \cite{yang2019condconv} and Dynamic Convolution (DyConv) \cite{chen2020dynamic} largely enhance the model capability by applying attention over layer-wise kernels. 
CondConv and DyConv bring considerable overheads of memory. There are three parts of memory overheads. 1) Since they use a linear combination of n kernels, the parameters are n times of static kernel. 2) The weights and the hidden layers of the auxiliary attention network bring extra memory overheads. 3) Because they dynamically generate the kernel for each input, there are $m$ kernels for $m$ inputs. As the batch size increases, CondConv and DyConv bring more memory overheads. 
Moreover, CondConv and DyConv cannot use existing convolution libraries directly when the batch size is larger than one, since existing libraries are designed for static convolution.

In this paper, we propose a new attention method, called Shift-and-Balance (SB). Shift-and-Balance also uses a light-weight attention branch that enhances the network's representational power by modeling channel-wise relationships. Different from Squeeze-and-Excite, the attention branch is gated using Tanh function and scaled by the learned control factor $\lambda$, then added into the feature map's trunk branch. The learned parameter $\lambda$ controls the attention branch's influence to coordinate and balance the trunk and attention branches' contributions.
Shift-and-Balance attention avoids the situation where the whole channel is inactivated.
Moreover, it avoids the problem of gradient vanishing at training time.
Shift-and-Balance can be effectively applied in layer-wise convolution to enhance the representational capability of each layer.
The overheads of Shift-and-Balance are close to Squeeze-and-Excite. The only difference is the integration of the attention and trunk branches. In Squeeze-and-Excite, the integration cost lies mainly in broadcasting multiplication, while ours lies mainly in broadcasting addition. In modern computers, addition is more efficient than multiplication. Moreover, our method can be easily implemented by using existing libraries directly. 

We evaluate Shift-and-Balance on the datasets of ImageNet, PASCAL VOC, and CIFAR-10. Experiments show that Shift-and-Balance significantly improves the accuracy compared to Squeeze-and-Excite when applied in more layers. Applying Squeeze-and-Excite in more layers degrades the accuracy dramatically on VOC and CIFAR-10 datasets. Applying in more layers allows Shift-and-Balance to increase more size and capacity of a network. Moreover, Shift-and-Balance achieves better accuracy than the state-of-art attention, DyConv, on the smaller network or smaller dataset. For example, Shift-and-Balance in MobileNetV2\_x0.35 and MobileNetV2\_x0.5 outperforms DyConv on ImageNet dataset, and Shift-and-Balance in MobileNetV2 of all scales outperforms DyConv on CIFAR-10 dataset.

\section{Related Work}


Squeeze-and-Excite (SE) \cite{Hu2018SqueezeandExcitationN} comprises a lightweight gating mechanism that enhances the network's representational power by modeling channel-wise relationships.
Residual attention network \cite{Wang2017ResidualAN} proposes a mixed attention mechanism that generates attention-aware features. Inside each attention module, an hourglass architecture is introduced to achieve global attention across both spatial and channel dimensions.
BAM \cite{Park2018BAMBA}, CBAM \cite{Woo2018CBAMCB}, scSE \cite{Roy2018ConcurrentSA} introduce spatial attention in addition to channel attention in a similar way.
Selective Kernel (SK) \cite{Li2019SelectiveKN} brings two branches with different kernel sizes, then fuses them using Softmax attention guided by the information of the two branches.
Split Attention \cite{Zhang2020ResNeStSN} enables feature-map attention across different feature-map groups. Attention to Scale \cite{Chen2016AttentionTS} proposes an attention mechanism that learns to softly weight the multi-scale features at each pixel location for image segmentation. Non-local neural network is proposed \cite{Wang2018NonlocalNN} for vision tasks such as video classification, object detection and instance segmentation based on the self-attention method \cite{Lin2017ASS} \cite{Vaswani2017AttentionIA}.
PSANet \cite{Zhao2018PSANetPS}, OCNet \cite{Yuan2018OCNetOC}, DANet \cite{Fu2019DualAN} also exploits the self-attention method for image segmentation. CCNet \cite{Huang2019CCNetCA} proposes criss-cross attention module to harvest the contextual information of all the pixels on its criss-cross path.

Recently, researchers successfully apply attention over weights instead of over features. Conditionally Parameterized Convolutions (CondConv) \cite{yang2019condconv} computes convolutional kernels as a function of the input instead of using static convolutional kernels. 
Dynamic Convolution (DyConv) \cite{chen2020dynamic} is a concurrent work with CondConv. The key idea is very similar to CondConv. The main difference is that it uses Softmax with a large temperature as the gate function of the kernel coefficient learned by the auxiliary network, while CondConv uses Sigmoid. 
WeightNet \cite{Ma2020WeightNetRT} unifies Squeeze-and-Excite and CondConv into the same framework on weight space. It generalizes the two methods by simply adding one more grouped fully-connected layer to the attention activation layer.

Dynamic Filter Network \cite{Jia2016DynamicFN} generates all the convolutional filters dynamically conditioned on the input, while CondConv and DyConv only generate the coefficient of a group of static filters. 
Dynamic ReLU \cite{Chen2020DynamicR} proposes a dynamic rectifier of which parameters are generated by a hyper-function over all input elements.
Other dynamic networks \cite{lin2017runtime} \cite{liu2018dynamic} \cite{wang2018skipnet} \cite{wu2018blockdrop} \cite{yu2018slimmable} \cite{huang2017multi} try to learn dynamic network structure with static convolution kernels.

\section{Problem of Existing Attention}
\subsection{Problem of Scaled Attention}
Squeeze-and-Excite (SE) \cite{Hu2018SqueezeandExcitationN} can be seen as scaled attention. Without loss of generality, scaled attention can be defined as
\begin{equation} \label{eq-se}
y = A(x)T(x) 
\end{equation}
where x is the input features, T(x) is the trunk branch. In original SE, $T(x)=x$. Extendedly, it can consist of one or more convolutional layers. A(x) is the attention branch defined as
\begin{equation} \label{eq-se2}
A(x) = Sigmoid(F((x)) 
\end{equation}
where F is a light-weight network, consisting of global average pooling (GAP) and a multi-layer fully-connected network. Sigmoid is the gate function. The output of A(x) is channel-wise and multiplied by T(x) with broadcasting along spatial dimensions.

It is hard to coordinate and balance the trunk and attention branches' contributions. 
In Equation \ref{eq-se}, the trunk branch is scaled with the coefficient computed by the attention branch. The input of the attention branch is the global average of the x. There is more information loss than the trunk branch, which takes the original x as the input. As a result, the attention branch should play a smaller role than the trunk branch in the final output. The gate function Sigmoid constrains the attention branch to a certain degree. However, scaled attention makes the output vary in a large range, from 0 to T(x). The influence of attention is still too significant. In other words, scaled attention is too sensitive to coordinate and balance the two branches' contributions. 
In the extreme situation, it makes the whole channel inactivated when Sigmoid is saturated on the side of zero. Inactivated channel makes a waste of representational capability, especially for the light-weight network with few channels. Moreover, each channel in a convolutional layer requires additional multiply-adds. The inactivated channel leads to a waste of computation.

Scaled attention using Sigmoid also hinders the gradient backpropagation of the loss $\ell$ at training time.

\begin{eqnarray}\label{eq-se-bp}
\frac{\partial \ell}{\partial x} & = & \frac{\partial \ell}{\partial y} \frac{\partial y}{\partial x} 
= \frac{\partial \ell}{\partial y} \frac{\partial (A(x)T(x))}{\partial x}
\nonumber\\
& = & \frac{\partial \ell}{\partial y} (\frac{\partial A(x)}{\partial x} T(x) + A(x)\frac{\partial T(x)}{\partial x} )
\end{eqnarray}

For simplicity, we use S to denote Sigmoid, and z to denote F(x). Because of Equation \ref{eq-se2}, we have
\begin{equation} \label{eq-se-bp2}
\frac{\partial A(x)}{\partial x} T(x) = S(z)(1-S(z))\frac{\partial z}{\partial x}T(x)
\end{equation}

\begin{equation} \label{eq-se-bp3}
A(x)\frac{\partial T(x)}{\partial x} = S(z)\frac{\partial T(x)}{\partial x}
\end{equation}

As shown in Equation \ref{eq-se-bp}, the gradient of loss $\frac{\partial \ell}{\partial x}$ depends on the sum of Equation \ref{eq-se-bp2} and Equation \ref{eq-se-bp3}.
In Equation \ref{eq-se-bp2}, the gradient would vanish when Sigmoid is saturated on both sides of zero and one. In Equation \ref{eq-se-bp3}, the gradient would vanish when Sigmoid is saturated on the side of zero. As a result, the final gradient would vanish when Sigmoid is saturated on the side of zero.

Using Tanh instead of Sigmoid can alleviate the inactivation of channel and gradient vanishing. However, the results vary between wider ranges, from $-T(x)$ to $T(x)$. That aggravates the sensitivity of the model. 

\subsection{Problem of Weight Attention}

CondConv \cite{yang2019condconv} largely enhances the model capability by applying attention over layer-wise kernels. In particular, the convolutional kernels in a CondConv layer are over-parameterized as a linear combination of n kernels $(\pi_1(x)W_1+...+\pi_n(x)W_n)*x$, where $\pi_1(x),...,\pi_n(x)$ are attention functions of the input learned through a light auxiliary network. 
CondConv would inactivate the candidate kernel but would not inactivate the channel of the feature. 
Dynamic Convolution (DyConv) \cite{chen2020dynamic} is a concurrent work with CondConv. The key idea is very similar to CondConv. The main difference is that it uses Softmax with a large temperature as the gate function of the kernel coefficient. Softmax with a large temperature restricts the attention output, thus facilitates the learning of the attention model.

CondConv and DyConv bring considerable memory overheads. There are three parts of memory overheads. 1) Since they use a linear combination of n kernels, the parameters are n times of static kernel. 2) The weights and the hidden layers of the auxiliary network bring extra memory overheads. 3) Because they dynamically generate the kernel for each input, there are $m$ kernels for $m$ inputs. As the batch size increases, CondConv and DyConv bring more memory overheads. 
Concerning computational overheads, the extra costs of CondConv and DyConv are only caused by the auxiliary network. However, CondConv and DyConv cannot use existing convolution libraries directly when the batch size is larger than one, since the existing libraries are designed for static convolution.

\section{Shift-and-Balance Attention}

\begin{figure}[tbh!]
\centering
\includegraphics[width=0.45\columnwidth]{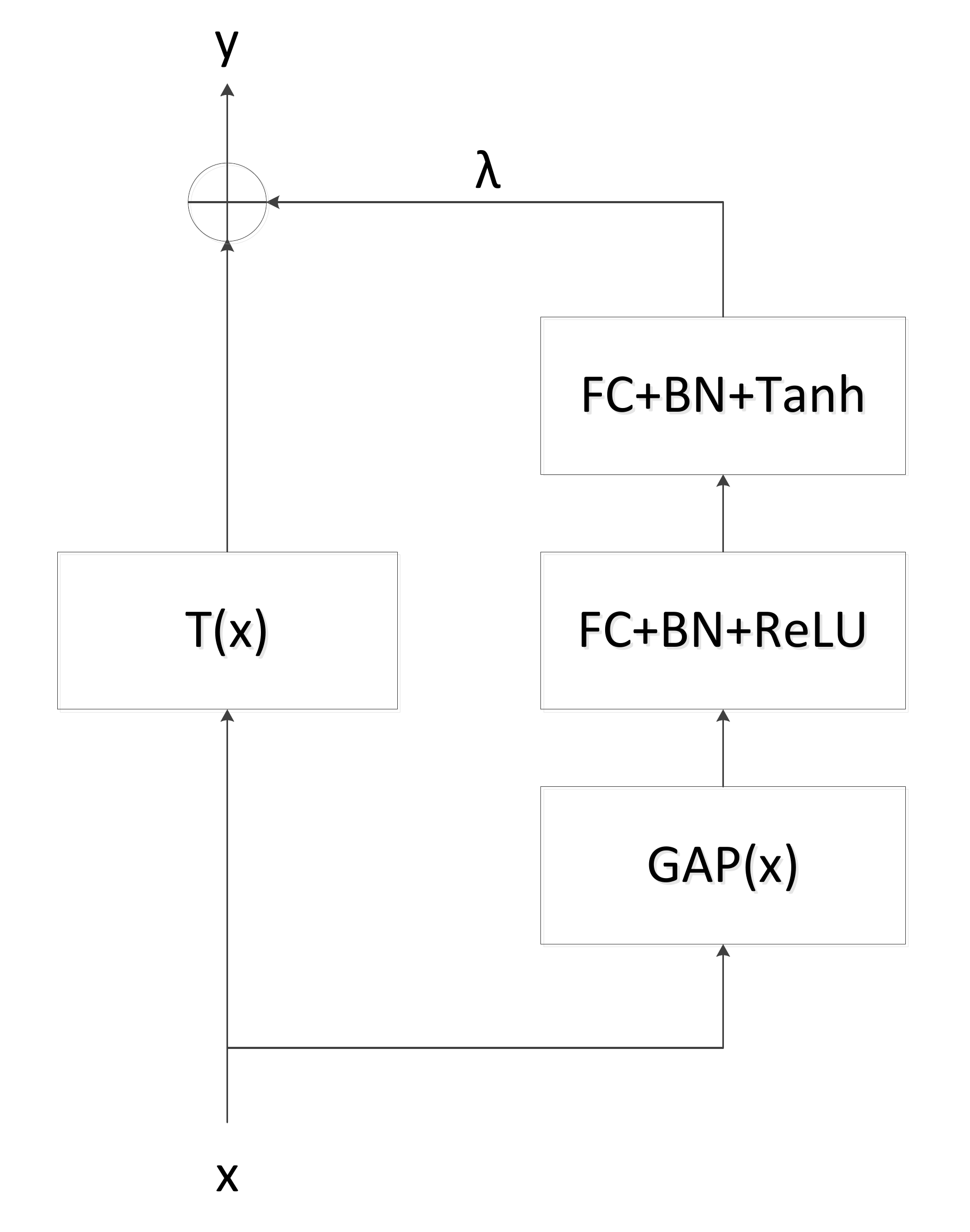}
\caption{ Shift-and-Balance attention. GAP refers to global average pooling, FC refers to fully-connected layer, BN refers to batch normalization. T(x) is the trunk branch consisting of one or more convolutional layers. $\oplus$ refers to channel-wise addition with broadcasting along spatial dimensions. $\lambda$ is the learned control factor to coordinate and balance the trunk and attention branches.}
\label{fig-sse}
\end{figure}

In this paper, we propose a new attention method, called Shift-and-Balance, which is defined as

\begin{equation} \label{eq-sa}
y = T(x) + \lambda A(x)
\end{equation}
where T(x) is the trunk branch consisting of one or more convolutional layers. $\lambda$ is the control factor learned automatically, and A(x) is defined as
\begin{equation} \label{eq-sa2}
A(x) = Tanh(F(x)) 
\end{equation}
As shown in Figure \ref{fig-sse}, F is a light-weight network consisting of global average pooling (GAP) and a multi-layer fully-connected network. We use Tanh by default as the gate function considering the symmetry of shift. The output of A(x) is channel-wise and multiplied by the channel-wise parameter $\lambda$, then added to T(x) with broadcasting along spatial dimensions.

\begin{figure}[tbh!]
\centering
\subfloat[scale]{
\label{fig-scale}
\includegraphics[width=0.45\columnwidth]{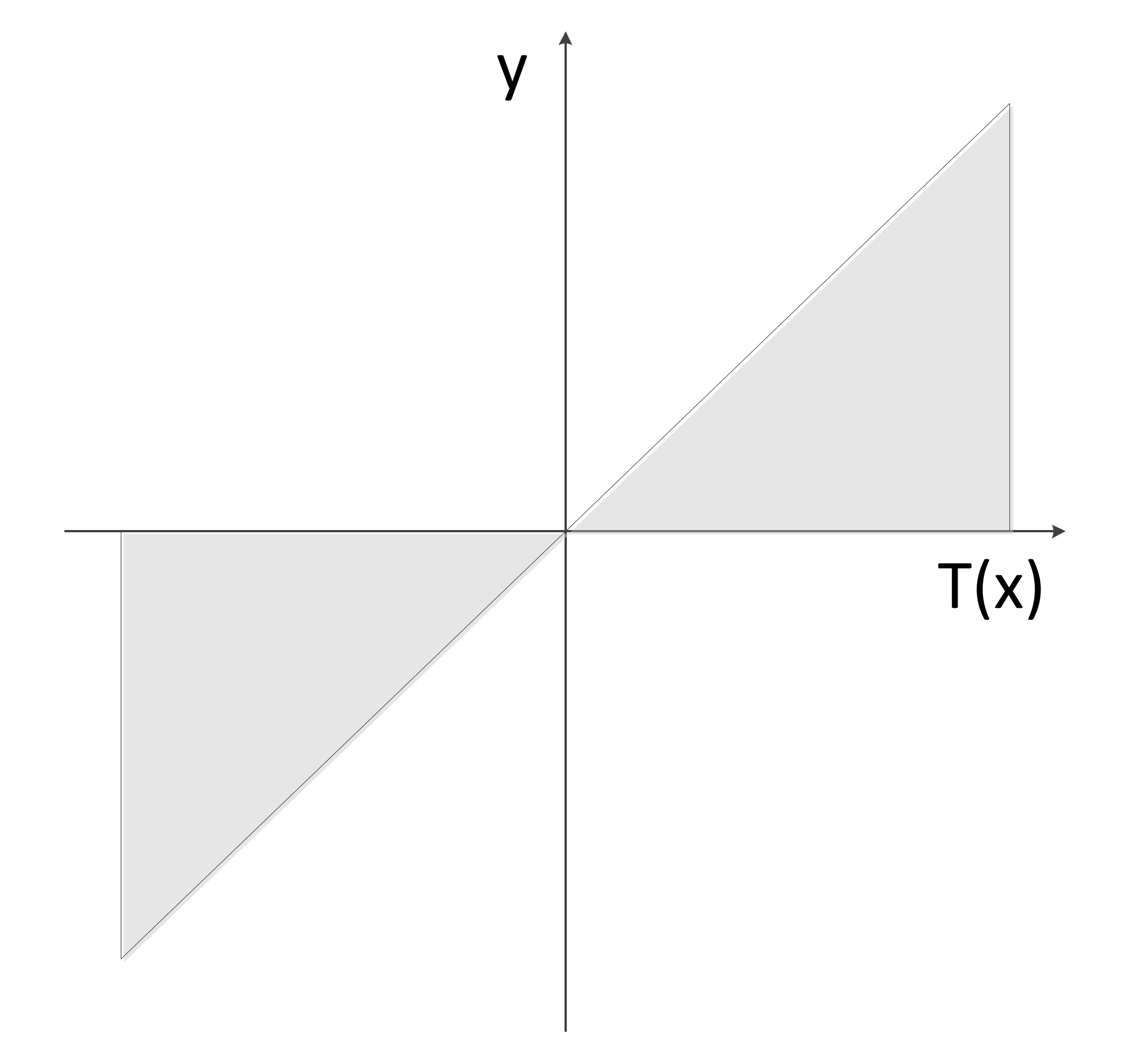}} 
\centering
\subfloat[shift ]{
\label{fig-shift}
\includegraphics[width=0.45\columnwidth]{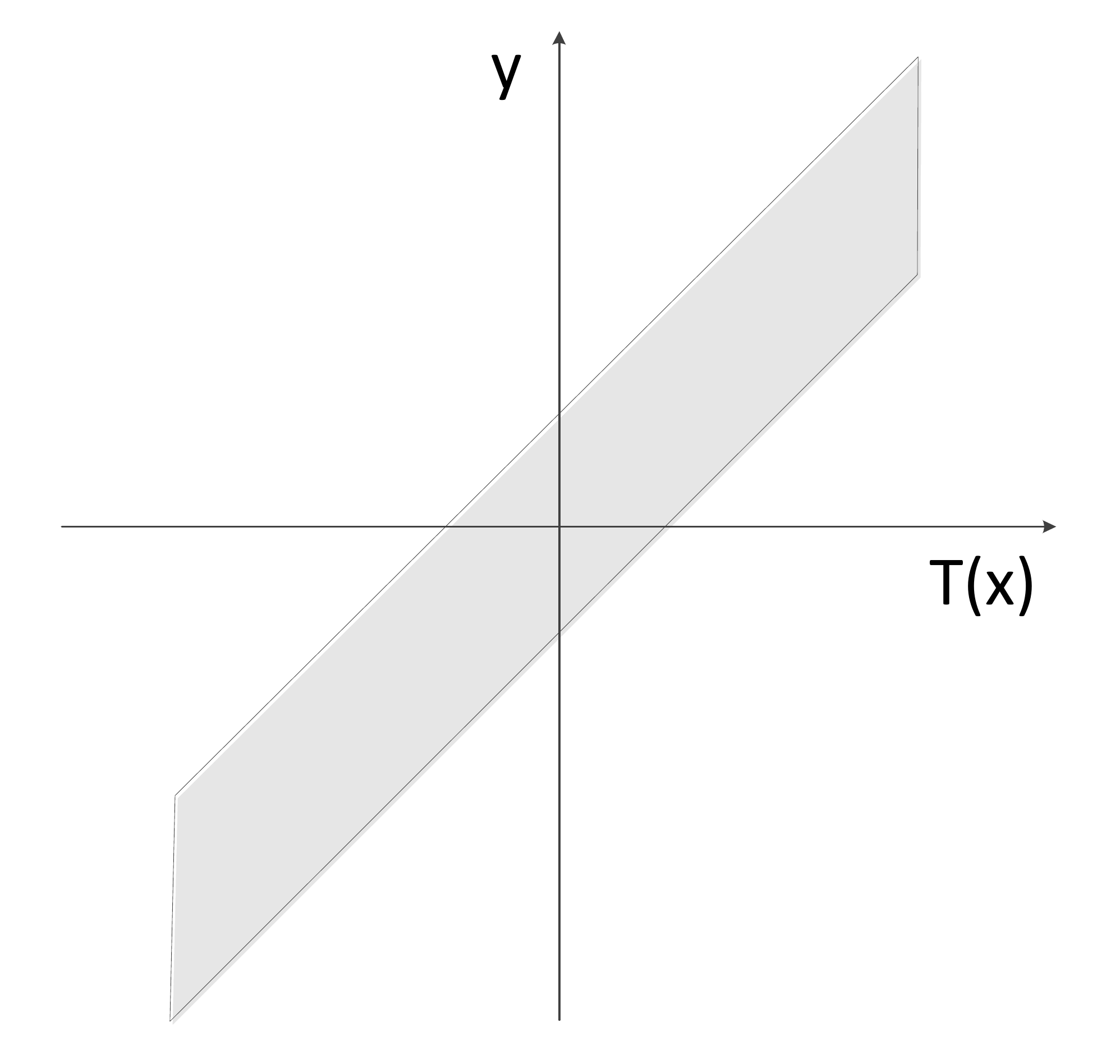}} 
\caption{ The difference of the output range between scaled attention and shifted attention. Applying attention over T(x), the output of scaled attention ranges from 0 to T(x), while the output of shifted attention ranges from $(T(x)-\lambda)$ to $(T(x)+\lambda)$.}
\label{fig-attention-diff}
\end{figure}

Shift-and-Balance attention alleviates the problem of scaled attention. The trunk branch is added by the attention, rather than multiplied. The influence degree of the attention branch can be controlled by the learned parameter $\lambda$. The shift range is between $-\lambda$ and $\lambda$. Thus the output of Shift-and-Balance ranges from $(T(x)-\lambda)$ to $(T(x)+\lambda)$. $\lambda$ is learned automatically from the data by backpropagation to coordinate and balance the trunk and attention branches. Figure \ref{fig-attention-diff} demonstrates the difference in the output range between scaled attention and shifted attention. Shift-and-Balance also can avoid the situation where the whole channel is inactivated when Sigmoid is saturated on the side of zero.

Moreover, Shift-and-Balance avoids the problem of gradient vanishing at training time. For Shift-and-Balance, the gradient of loss $\ell$ is computed as 

\begin{eqnarray}\label{eq-sse-bp}
\frac{\partial \ell}{\partial x} & = & \frac{\partial \ell}{\partial y} \frac{\partial y}{\partial x} 
= \frac{\partial \ell}{\partial y} \frac{\partial (T(x)+\lambda A(x))}{\partial x}
\nonumber\\
& = & \frac{\partial \ell}{\partial y} (\frac{\partial T(x)}{\partial x} + \lambda \frac{\partial A(x)}{\partial x} )
\end{eqnarray}

For simplicity, we also use z to denote F(x). Because of Equation \ref{eq-sa2}, we have
\begin{equation} \label{eq-sse-bp2}
\frac{\partial A(x)}{\partial x} = (1-Tanh^2(z))\frac{\partial z}{\partial x}
\end{equation}

We can see that the attention branch does not affect $\frac{\partial T(x)}{\partial x}$. 
That is to say, there is no gradient vanishing for the trunk branch. Thus the gradient backpropagation through the trunk branch is unhindered.

Actually, the balance of the two branches depends on the addition mechanism and the control factor $\lambda$, rather than the gate function. Although we use Tanh as the default gate function, other functions are also effective, e.g., Sigmoid. In Shift-and-Balance, Sigmoid saturated on the side of zero does not inactivate the trunk branch's channel. It only inhibits the attention branch and makes the attention branch sparse.

\subsection{Overheads}

The extra overheads of our method are caused by the auxiliary network. 
Take the layer-wise attention as example, where the input feature size is $c_{in} \times h \times w$, the output feature size is $c_{out} \times h \times w$, T(x) takes a standard convolution layer with kernel size of $c_{in} \times c_{out} \times k \times k$, and the hidden units of the fully-connected network is $c_{hid}$. To reflect the computational cost, we use Multiplies and Adds to refer to the number of multiplies and additions.

The computational overheads of Shift-and-Balance consist of three parts, the GAP, the two fully-connected layers, and the integration of T(x) and A(x). 
We ignore the normalization and the activation function because it either can be fused at inference time or takes little cost.
The GAP sums the features per channel and is divided by the number of features per channel. It has the cost  
\begin{equation} \label{eq-cost3}
Multiplies = c_{in} 
\end{equation}
\begin{equation} \label{eq-cost4}
Adds = c_{in} \times h \times w
\end{equation}
The two fully-connected layers have the cost  
\begin{equation} \label{eq-cost5}
Multiplies = c_{in} \times c_{hid} + c_{hid} \times c_{out}
\end{equation}
\begin{equation} \label{eq-cost6}
Adds = c_{in} \times c_{hid} + c_{hid} \times c_{out}
\end{equation}
The integration of $T(x)$ and $A(x)$ consists of multiplication $A(x)$ by $\lambda$, and broadcasting addition between $T(x)$ and $\lambda A(x)$. It has the cost  
\begin{equation} \label{eq-cost7}
Multiplies = c_{out} 
\end{equation}
\begin{equation} \label{eq-cost8}
Adds = c_{out} \times h \times w
\end{equation}
As a result, the total extra costs of our attention are 
\begin{eqnarray}\label{eq-cost9}
Multiplies & = & c_{in} + c_{in} \times c_{hid} 
\nonumber\\
& & + c_{hid} \times c_{out} + c_{out} 
\end{eqnarray}
\begin{eqnarray}\label{eq-cost10}
Adds & = & c_{in} \times h \times w + c_{in} \times c_{hid} 
\nonumber\\
& & + c_{hid} \times c_{out} + c_{out} \times h \times w
\end{eqnarray}

The overheads of our method are close to Squeeze-and-Excite. The only difference is the integration of $T(x)$ and $A(x)$. In Squeeze-and-Excite, the integration cost lies mainly in broadcasting multiplication, while ours lies mainly in broadcasting addition. In modern computers, addition is more efficient than multiplication.
Moreover, our method can be easily implemented by using existing libraries directly. 


\section{Evaluation}

\subsection{ImageNet} \label{sec-imagenet}

\begin{table*}[tbh!]
\begin{center}
\begin{tabular}{l|l|c|c|c}
\hline
& & Parameters & MAdds & Accuracy(\%) \\
\hline
\multirow{4}{*}{MobileNetV2\_x0.35}& static & 1.677M & 59.2M & 57.826\\
& SE & 1.736M & 59.5M & 59.106\\
& DyConv & 2.690M & 62.0M & 62.136\\
& SB, ours & 2.700M & 60.2M & \textbf{62.290}\\
\hline
\multirow{4}{*}{MobileNetV2\_x0.5}& static & 1.969M & 97.0M & 62.712\\
& SE & 2.086M & 97.4M & 64.964\\
& DyConv & 3.951M & 101.4M & 66.754\\
& SB, ours & 4.006M & 99.0M & \textbf{67.066}\\
\hline
\multirow{4}{*}{MobileNetV2\_x0.75} & static & 2.636M & 209.0M & 68.278\\
& SE & 2.898M & 209.8M & 70.104\\
& DyConv & 7.018M & 217.5M & \textbf{71.392}\\
& SB, ours & 7.201M & 213.5M & 71.044\\
\hline
\multirow{4}{*}{MobileNetV2\_x1.0} & static & 3.505M & 300.0M & 70.806\\
& SE & 3.966M & 301.0M & 72.426\\
& DyConv & 11.158M & 312.9M & \textbf{73.348}\\
& SB, ours & 11.530M & 308.0M & 72.694\\

\hline
\end{tabular}
\end{center}
\caption{Top-1 accuracies of MobileNetV2 with different width multipliers on ImageNet validation dataset. }
\label{tab-imagenet}
\end{table*}

ImageNet classification dataset \cite{russakovsky2015imagenet} has 1.28M training images and 50,000 validation images with 1000 classes. We use Pytorch in our experiments, and we use the same procedure as the official examples of Pytorch \footnote{https://github.com/pytorch/examples/tree/master/imagenet}. To augment data, we crop the training images with the random size of 0.08 to 1.0 and a random aspect ratio of 3/4 to 4/3, and then resize to 224x224.  Then random horizontal flipping is made. The validation image is resized to 256x256, and then cropped by 224x224 at the center. Each channel of the input is normalized into 0 mean and 1 std globally. SGD with momentum 0.9 and batch size 256 is used for training. 
All the settings above are the same as the official examples of Pytorch, except that we use weight decay of 4e-5 instead of 1e-4 according to the common setting for training light-weight networks. 
We train the networks with 300 epochs using linear-decay learning rate policy, decreased from 0.1 to 0. 
Four TITAN Xp GPUs are used to train the networks.

We first evaluate MobileNetV2 with different width multipliers. 
We compare static convolution, Squeeze-and-Excite (SE) \cite{Hu2018SqueezeandExcitationN}, dynamic convolution (DyConv) \cite{chen2020dynamic}, and our Shift-and-Balance (SB).
Following the original paper \cite{Hu2018SqueezeandExcitationN}, we apply SE in the bottlenecks' output. DyConv and SB are used in all the convolutional layers in the inverted bottlenecks.
For DyConv, the number of experts $n$ is set to the default value of 4 \cite{chen2020dynamic}. 
The auxiliary attention network is a two-layer fully-connected network with the hidden units to be $1/4$ input channels. As recommended in the original paper \cite{chen2020dynamic}, we use Softmax with temperature 30 as the gate function of the auxiliary network. For DyConv, the additional parameters mainly lie in the extra experts rather than the auxiliary attention network. 
For SE and SB, the additional parameters lie in the auxiliary attention network. For SE, the hidden units are set to be equal to output channels. For SB, we set the auxiliary network's hidden units in pointwise convolutional layers equal to the output channels. Because the depth-wise convolutional layer is expanded by 6 times in the inverted bottlenecks, we set the hidden units to be $1/6$ output channels in the depthwise convolutional layer to reduce parameters. The $\lambda$ in SB is initialized with 0.1.
Dropout \cite{Srivastava2014DropoutAS} with 0.2 is used in the last fully connected layers of MobileNetV2. All the networks are trained by ourselves using the same training settings.

The results are shown in Table \ref{tab-imagenet}. We can see that all attentions improve the accuracy compared to the static model without attention. SB outperforms the original SE with all width multipliers. Moreover, SB achieves better accuracies than DyConv in MobileNetV2\_x0.35 and MobileNetV2\_x0.5 with close parameters and MAdds(Multiply-Adds). As the width increases, SB achieves lower accuracy than DyConv. We can also see that SE achieves decent accuracy in MobileNetV2\_x1.0 with much fewer parameters. That implies the effect of over-parameterization is more pronounced in smaller networks. With more parameters, we need sufficient regularization to avoid overfitting. 

The original SE is applied block-wise in the output of the bottlenecks. We further apply SE in layer-wise convolution to compare with SB fairly. We first use SE and SB only in the pointwise convolutional layers of the bottlenecks. Then use them in more layers, both the pointwise and depthwise convolutional layers. Table \ref{tab-scale-shift} shows the results. We find both SE and SB outperform the baseline. Moreover, for both SE and SB, the accuracy increases as the attention is used in more layers on the large-scale ImageNet dataset. We can also find SB achieves higher accuracy than SE with little extra parameters caused by the control factor $\lambda$.

\begin{table}[tbh!]
\begin{center}
\begin{tabular}{l|c|c}
\hline
&Paras & Acc(\%) \\
\hline
baseline, static & 1.677 &57.826 \\
\hline
SE, Pointwise & 2.500 & 59.960\\
SB, Pointwise & 2.503 & \textbf{62.010}\\
\hline
SE, Pointwise + Depthwise & 2.695 & 60.862 \\
SB, Pointwise + Depthwise & 2.700 & \textbf{62.290}\\

\hline
\end{tabular}
\end{center}
\caption{Comparison of Squeeze-and-Excite (SE) and Shift-and-Balance (SB) using in more layers. The baseline is static MobileNetV2\_x0.35 without attention. Attention is applied in 1) the pointwise convolutional layers, 2) both the pointwise and depthwise convolutional layers. Paras is short for parameters (millions). Acc is short for accuracy. }
\label{tab-scale-shift}
\end{table}

We next evaluate the effectiveness of SB on other network architectures. We train ShuffleNetV2\_x0.5 and MnasNet\_x0.5 using the same training settings as MobileNetV2, except the hidden units are set to be equal to output channels in the depthwise convolutional layer in ShuffleNetV2. Both SE and SB are applied in all three layers of the ShuffleNetV2 and MnasNet blocks. From Table \ref{tab-shuffle-mnas}, we can find that SB improves the accuracies of ShullfeNetV2 and MnasNet, while SE decreases the accuracies compared to the static network without attention.

\begin{table}[tbh!]
\begin{center}
\begin{tabular}{l|l|c|c}
\hline
& & Paras & Acc(\%) \\
\hline
\multirow{2}{*}{ShuffleNetV2\_x0.5} & static & 1.367 &59.068\\
& SE & 1.772 & 58.482\\
& SB & 1.775 & \textbf{62.314}\\
\hline
\multirow{2}{*}{MnasNet\_x0.5} & static & 2.219 & 65.336 \\
& SE & 5.224 & 63.720\\
& SB & 5.233 & \textbf{68.114}\\

\hline
\end{tabular}
\end{center}
\caption{Top-1 accuracies of ShuffleNetV2 and MnasNet on ImageNet validation dataset. Paras is short for parameters (millions). Acc is short for accuracy. Both SE and SB are applied in all three layers of the ShuffleNetV2 and MnasNet blocks.}
\label{tab-shuffle-mnas}
\end{table}

\subsection{VOC}

We evaluate one-stage object detection using Single Shot MultiBox Detector (SSD), training on the PASCAL VOC2007 + 2012 training and validation sets, and reporting the mean average precision (mAP) on the PASCAL VOC2007 test set. 
VOC2007 + 2012 training and validation sets have 5011 + 11540 images, and VOC2007 test set has 4952 images. There are 20 classes of objects to be detected.
Our code is based on the project of pytorch-ssd \footnote{https://github.com/qfgaohao/pytorch-ssd}.

MobileNetV2 with different width multipliers is used as the backbone. 
All the networks are trained using SGD with momentum 0.9 and weight decay 4e-5. The batch size is set to 32. The learning rate is set to 0.01 and scheduled to arrive at zero using the cosine annealing scheduler. We train the networks with 200 epochs. 
SE and SB are applied in all the convolutional layers in the backbone's inverted bottlenecks. The settings of SE and SB are the same as Section \ref{sec-imagenet}.

The results are shown in Table \ref{tab-voc}. We find SB improves the performance compared to static SSD-MobileNetV2 with different width multipliers. We can also find that applying SE in all the convolutional layers degrades the accuracy dramatically on VOC dataset. SE even corrupts the model when training SSD-MobileNetV2\_x0.5.

\begin{table}[tbh!]
\begin{center}
\begin{tabular}{l|l|c}
\hline
& & mAP(\%) \\
\hline
\multirow{2}{*}{SSD-MobileNetV2\_x0.35} & static & 45.409\\
& SE & 11.025\\
& SB & \textbf{46.795}\\
\hline
\multirow{2}{*}{SSD-MobileNetV2\_x0.5} & static & 51.770\\
& SE & / \\
& SB & \textbf{52.245}\\
\hline
\multirow{2}{*}{SSD-MobileNetV2\_x0.75} & static & 56.304\\
& SE & 14.125\\
& SB & \textbf{57.325}\\
\hline
\multirow{2}{*}{SSD-MobileNetV2\_x1.0} & static & 59.966\\
& SE & 21.168\\
& SB & \textbf{61.491}\\

\hline
\end{tabular}
\end{center}
\caption{The mean average precision (mAP) on the PASCAL VOC2007 test set. SE corrupts the model when training SSD-MobileNetV2\_x0.5}
\label{tab-voc}
\end{table}

\subsection{CIFAR} \label{sec-cifar}

\begin{table*}[tbh!]
\begin{center}
\begin{tabular}{l|ccc|cc|cc}
\hline
& \multirow{2}{*}{ C1} & \multirow{2}{*}{C2} & \multirow{2}{*}{C3} & \multicolumn{2}{c|}{ SE} & \multicolumn{2}{c}{SB} \\
\cline{5-8}
& & & & Parameters & Accuracy(\%) & Parameters & Accuracy(\%)\\
\hline
\multirow{3}{*}{MobileNetV2\_x0.35} & \checkmark & - & - & 1.089M & $\textbf{90.980}_{\pm 0.191}$ & 1.091M & $90.826_{\pm 0.117}$\\
& \checkmark & - & \checkmark & 1.240M & $90.040_{\pm 0.518}$ & 1.243M & $\textbf{91.252}_{\pm 0.255}$\\
& \checkmark & \checkmark & \checkmark & 1.435M & $84.184_{\pm 2.583}$ & 1.441M & $91.220_{\pm 0.137}$\\
\hline
\multirow{3}{*}{MobileNetV2\_x0.5} & \checkmark & - & - & 2.056M & $\textbf{91.582}_{\pm 0.368}$ & 2.060M & $91.340_{\pm 0.123}$\\
& \checkmark & - & \checkmark & 2.358M & $90.624_{\pm 0.382}$ & 2.363M & $91.700_{\pm 0.167}$\\
& \checkmark & \checkmark & \checkmark & 2.745M & $84.476_{\pm 1.697}$ &2.753M & $\textbf{91.746}_{\pm 0.151}$\\
\hline
\multirow{3}{*}{MobileNetV2\_x0.75} & \checkmark & - & - & 4.410M & $\textbf{92.812}_{\pm 0.106}$ & 4.415M & $92.496_{\pm 0.139}$\\
& \checkmark & - & \checkmark & 5.088M & $91.484_{\pm 0.342}$ & 5.095M & $92.452_{\pm 0.238}$\\
& \checkmark & \checkmark & \checkmark & 5.954M & $87.380_{\pm 1.879}$ & 5.966M & $\textbf{92.590}_{\pm 0.073}$ \\
\hline
\multirow{3}{*}{MobileNetV2\_x1.0} & \checkmark & - & - & 7.587M & $\textbf{93.172}_{\pm 0.185}$ & 7.594M & $92.746_{\pm 0.212}$\\
& \checkmark & - & \checkmark & 8.783M & $92.508_{\pm 0.216}$ & 8.792M & $92.870_{\pm 0.115}$\\
& \checkmark & \checkmark & \checkmark & 10.304M & $86.784_{\pm 1.797}$ & 10.320M & $\textbf{92.972}_{\pm 0.142}$\\

\hline
\end{tabular}
\end{center}
\caption{Comparison between Squeeze-and-Excite (SE) and Shift-and-Balance (SB) at different layers on CIFAR-10. C1, C2, and C3 indicate the 1x1 pointwise convolution that expands output channels, the 3x3 depthwise convolution, and the 1x1 pointwise convolution that shrinks output channels per block, respectively. Tick (\checkmark) indicates applying attention in this layer, while hyphen(-) indicates applying no attention.}
\label{tab-cifar-layer}
\end{table*}

\begin{table}[tbh!]
\begin{center}
\begin{tabular}{l|l|c}
\hline
& & Accuracy(\%) \\
\hline
\multirow{3}{*}{MobileNetV2\_x0.35}& static & $90.854_{\pm 0.155}$\\
& DyConv & $90.974_{\pm 0.268}$\\
& SB & $\textbf{91.220}_{\pm 0.137}$\\
\hline
\multirow{3}{*}{MobileNetV2\_x0.5}& static & $91.240_{\pm 0.268}$\\
& DyConv & $91.614_{\pm 0.249}$\\
& SB & $\textbf{91.746}_{\pm 0.151}$\\
\hline
\multirow{3}{*}{MobileNetV2\_x0.75}& static & $92.228_{\pm 0.146}$\\
& DyConv & $92.506_{\pm 0.102}$\\
& SB & $\textbf{92.590}_{\pm 0.082}$\\
\hline
\multirow{3}{*}{MobileNetV2\_x1.0}& static & $92.410_{\pm 0.084}$\\
& DyConv & $92.694_{\pm 0.100}$\\
& SB & $\textbf{92.920}_{\pm 0.156}$\\

\hline
\end{tabular}
\end{center}
\caption{Accuracies of MobileNetV2 with different width multipliers on CIFAR-10.}
\label{tab-acc-cifar}
\end{table}

CIFAR-10 \cite{krizhevsky2009learning} is a dataset of natural 32x32 RGB images in 10 classes with 50, 000 images for training and 10, 000 for testing. To augment data, we pad the training images with 0 to 36x36 and then randomly crop them to 32x32 pixels. Then we carry out randomly horizontal flipping. Each channel of the input is normalized into 0 mean and 1 std globally.
We use SGD with momentum 0.9 and weight decay 5e-4. The batch size is set to 128. The learning rate is set to 0.1 and scheduled to arrive at zero using the cosine annealing scheduler. We train the networks with 200 epochs. All the settings above are the same as the project of pytorch-cifar \footnote{https://github.com/kuangliu/pytorch-cifar}. We run each test 5 times, and report the mean and the standard deviation of the accuracies.

We first evaluate static convolution, DyConv, and Shift-and-Balance (SB) on MobileNetV2 with different width multipliers. The attention settings are the same as Section \ref{sec-imagenet}. 
Table \ref{tab-acc-cifar} shows that SB increases the accuracy compared to the static MobileNetV2 without attention. Moreover, it outperforms DyConv on MobileNetV2 with different width multipliers.

We further compare SE and SB at different layers. Table \ref{tab-cifar-layer} shows the classification accuracy for using SE and SB at three different layers (1x1 pointwise convolution, 3x3 depthwise convolution, 1x1 pointwise convolution) per bottleneck block in MobileNetV2. Generally, the accuracy increases as SB is used in more layers, while the accuracy decreases as SE is used in more layers. Using SE in all three layers dramatically degrades the accuracy. However, we also notice that SE is better than SB when used in only one layer. In the larger networks, MobileNetV2\_x0.75 and MobileNetV2\_x1.0, SE used in one layer even outperforms SB used in all three layers. Because of more parameters, SB used in all three layers increases the risk of overfitting, especially on a small dataset. To add regularization and avoid overfitting, we use Dropout \cite{Srivastava2014DropoutAS} with 0.2 in the attention branch's hidden layer. Table \ref{tab-dropout} shows that Dropout improves the accuracy. With sufficient regularization, using SB in more layers can increase the network's size and capacity.

\begin{table}[tbh!]
\begin{center}
\begin{tabular}{l|c}
\hline
& Accuracy(\%) \\
\hline
SE, w/o Dropout & $86.784_{\pm 1.797}$\\
SE, w/ Dropout & $87.966_{\pm 1.711}$\\
SB, w/o Dropout & $92.972_{\pm 0.142}$\\
SB, w/ Dropout & $\textbf{93.216}_{\pm 0.138}$\\

\hline
\end{tabular}
\end{center}
\caption{Effect of Dropout on MobileNetV2\_x1.0.}
\label{tab-dropout}
\end{table}

\subsection{Ablation Study}

We perform a number of ablations with MobileNetV2 on CIFAR-10. Shift-and-Balance (SB) is applied in all the convolutional layers of the inverted bottlenecks. The default setups are the same as Section \ref{sec-cifar}. We run each test 5 times, and report the mean and the standard deviation of the accuracies.

\begin{table}[tbh!]
\begin{center}
\begin{tabular}{c|c|c}
\hline
&Initial $\lambda$ & Accuracy(\%) \\
\hline
\multirow{5}{*}{MobileNetV2\_x0.35} &0.01 & $91.288_{\pm 0.214}$ \\
&0.05 & $\textbf{91.440}_{\pm 0.358}$ \\
&0.1 & $91.220_{\pm 0.137}$ \\
&0.5 & $90.584_{\pm 0.474}$ \\
&1.0 & $89.084_{\pm 0.388}$ \\
\hline
\multirow{5}{*}{MobileNetV2\_x0.5} &0.01 & $\textbf{91.888}_{\pm 0.324}$ \\
&0.05 & $91.862_{\pm 0.168}$ \\
&0.1 & $91.746_{\pm 0.151}$ \\
&0.5 & $91.430_{\pm 0.064}$ \\
&1.0 & $90.310_{\pm 0.547}$ \\
\hline
\multirow{5}{*}{MobileNetV2\_x0.75} &0.01 & $\textbf{92.714}_{\pm 0.180}$ \\
&0.05 & $92.706_{\pm 0.160}$ \\
&0.1 & $92.590_{\pm 0.073}$ \\
&0.5 & $92.266_{\pm 0.163}$ \\
&1.0 & $91.330_{\pm 0.446}$ \\
\hline
\multirow{5}{*}{MobileNetV2\_x1.0} &0.01 & $92.976_{\pm 0.092}$ \\
&0.05 & $\textbf{93.054}_{\pm 0.249}$ \\
&0.1 & $92.972_{\pm 0.142}$ \\
&0.5 & $92.760_{\pm 0.202}$ \\
&1.0 & $92.310_{\pm 0.302}$ \\
\hline
\end{tabular}
\end{center}
\caption{Different initial values for the learned control factor $\lambda$.}
\label{tab-init}
\end{table}

\begin{table}[tbh!]
\begin{center}
\begin{tabular}{l|c|c}
\hline
& Gate & Accuracy(\%) \\
\hline
\multirow{5}{*}{MobileNetV2\_x0.35} & Tanh & $91.220_{\pm 0.137}$ \\
&Sigmoid & $\textbf{91.306}_{\pm 0.244}$ \\
&Softmax & $90.894_{\pm 0.435}$ \\
&Relu & $90.604_{\pm 0.728}$ \\
&No Gate & $90.762_{\pm 0.570}$ \\
\hline
\multirow{5}{*}{MobileNetV2\_x0.5} & Tanh & $\textbf{91.746}_{\pm 0.151}$\\
&Sigmoid & $91.728_{\pm 0.193}$\\
&Softmax & $91.482_{\pm 0.227}$\\
&Relu & $91.624_{\pm 0.148}$\\
&No Gate & $91.296_{\pm 0.681}$\\
\hline
\multirow{5}{*}{MobileNetV2\_x0.75} & Tanh & $92.590_{\pm 0.073}$\\
&Sigmoid & $92.466_{\pm 0.096}$\\
&Softmax & $92.136_{\pm 0.180}$\\
&Relu & $92.438_{\pm 0.096}$\\
&No Gate & $\textbf{92.654}_{\pm 0.164}$\\
\hline
\multirow{5}{*}{MobileNetV2\_x1.0} & Tanh & $92.972_{\pm 0.142}$\\
&Sigmoid & $92.854_{\pm 0.131}$\\
&Softmax & $92.494_{\pm 0.241}$\\
&Relu & $92.760_{\pm 0.171}$\\
&No Gate & $\textbf{92.988}_{\pm 0.075}$\\

\hline
\end{tabular}
\end{center}
\caption{Different gate functions. }
\label{tab-gate}
\end{table}

We first investigate different initial values for the learned control factor $\lambda$ in Table \ref{tab-init}. Although we use 0.1 as the initial value in the above experiments, we find using a smaller initial value can further increase the accuracy. All models achieve close performances with initial values 0.01, 0.05, and 0.1, better than with 0.5 and 1.0. MobileNetV2\_x0.35 and MobileNetV2\_x1.0 achieve the best performance using initial value 0.05, and MobileNetV2\_x0.5 and MobileNetV2\_x0.75 achieve the best performance using initial value 0.01.

We investigate different choices for the gate function of the attention branch in Table \ref{tab-gate}. Generally, it is safe to use Tanh since it achieves the best or close to the best accuracy.
Sigmoid achieves slightly lower accuracy than Tanh, except on MobileNetV2\_x0.35 where Sigmoid is the best. We find that SB without any gate function achieves the best accuracy on MobileNetV2\_x0.75 and MobileNetV2\_x1.0. That indicates the learned control factor $\lambda$ may be strong enough to regulate the attention. Omitting the gate function could further save the computation cost and improve efficiency. We also try different gate functions for SE. When applied in all the convolutional layers, SE works only with Sigmoid function. Other gate functions corrupt the model.

We next investigate different choices for the hidden units of the attention branch. We increase the hidden units by 1, 2, 3, 4 times the output channels on MobileNetV2\_x0.35 with SB. Table \ref{tab-hidden} shows that increasing the hidden units can further improve the accuracy for MobileNetV2\_x0.35. The model achieves the best performance with hidden units to be 3 times the output channels. Too many hidden units (4 times) make the model overfitting, thus decrease the accuracy.

\begin{table}[tbh!]
\begin{center}
\begin{tabular}{c|c|c}
\hline
Hidden layer & Parameters & Accuracy(\%) \\
\hline
x1 & 1.441M & $91.220_{\pm 0.137}$\\
x2 & 2.447M & $91.286_{\pm 0.265}$\\
x3 & 3.454M & $\textbf{91.562}_{\pm 0.122}$\\
x4 & 4.460M & $90.900_{\pm 0.416}$\\

\hline
\end{tabular}
\end{center}
\caption{Different hidden units of the auxiliary attention network.}
\label{tab-hidden}
\end{table}

\section{Conclusion}
In this paper, we propose a new attention method, called Shift-and-Balance (SB).
SB uses a light-weight attention branch that enhances the network's representational power by modeling channel-wise relationships.
Different from Squeeze-and-Excite (SE), the attention branch is regulated by the gate function and the learned control factor, then added into the feature map's trunk branch. 
Experiments show that SB significantly improves the accuracy compared to SE when applied in more layers, increasing more size and capacity of a network. Moreover, SB achieves better or close accuracy compared to DyConv, especially on the smaller network or smaller dataset.




{\small
\bibliographystyle{ieee_fullname}
\bibliography{ref}

\begin{thebibliography}{10}\itemsep=-1pt

\bibitem{Chen2016AttentionTS}
Liang-Chieh Chen, Y. Yang, Jiang Wang, Wei Xu, and A. Yuille.
\newblock Attention to scale: Scale-aware semantic image segmentation.
\newblock {\em 2016 IEEE Conference on Computer Vision and Pattern Recognition
  (CVPR)}, pages 3640--3649, 2016.

\bibitem{chen2020dynamic}
Yinpeng Chen, Xiyang Dai, Mengchen Liu, Dongdong Chen, Lu Yuan, and Zicheng
  Liu.
\newblock Dynamic convolution: Attention over convolution kernels.
\newblock In {\em Proceedings of the IEEE/CVF Conference on Computer Vision and
  Pattern Recognition}, pages 11030--11039, 2020.

\bibitem{Chen2020DynamicR}
Y. Chen, X. Dai, Mengchen Liu, Dongdong Chen, Lu Yuan, and Zicheng Liu.
\newblock Dynamic relu.
\newblock {\em ArXiv}, abs/2003.10027, 2020.

\bibitem{Fu2019DualAN}
J. Fu, J. Liu, Haijie Tian, Z. Fang, and H. Lu.
\newblock Dual attention network for scene segmentation.
\newblock {\em 2019 IEEE/CVF Conference on Computer Vision and Pattern
  Recognition (CVPR)}, pages 3141--3149, 2019.

\bibitem{He2016DeepRL}
Kaiming He, Xiangyu Zhang, Shaoqing Ren, and Jian Sun.
\newblock Deep residual learning for image recognition.
\newblock {\em 2016 IEEE Conference on Computer Vision and Pattern Recognition
  (CVPR)}, pages 770--778, 2016.

\bibitem{Howard2019SearchingFM}
Andrew Howard, Mark Sandler, Grace Chu, Liang-Chieh Chen, Bo Chen, Mingxing
  Tan, Weijun Wang, Yukun Zhu, Ruoming Pang, Vijay Vasudevan, Quoc~V. Le, and
  Hartwig Adam.
\newblock Searching for mobilenetv3.
\newblock {\em 2019 IEEE/CVF International Conference on Computer Vision
  (ICCV)}, pages 1314--1324, 2019.

\bibitem{Howard2017MobileNetsEC}
Andrew~G. Howard, Menglong Zhu, Bo Chen, Dmitry Kalenichenko, Weijun Wang,
  Tobias Weyand, Marco Andreetto, and Hartwig Adam.
\newblock Mobilenets: Efficient convolutional neural networks for mobile vision
  applications.
\newblock {\em ArXiv}, abs/1704.04861, 2017.

\bibitem{Hu2018SqueezeandExcitationN}
Jie Hu, Li Shen, and Gang Sun.
\newblock Squeeze-and-excitation networks.
\newblock {\em 2018 IEEE/CVF Conference on Computer Vision and Pattern
  Recognition}, pages 7132--7141, 2018.

\bibitem{huang2017multi}
Gao Huang, Danlu Chen, Tianhong Li, Felix Wu, Laurens van~der Maaten, and
  Kilian~Q Weinberger.
\newblock Multi-scale dense networks for resource efficient image
  classification.
\newblock {\em arXiv preprint arXiv:1703.09844}, 2017.

\bibitem{Huang2017DenselyCC}
Gao Huang, Zhuang Liu, and Kilian~Q. Weinberger.
\newblock Densely connected convolutional networks.
\newblock {\em 2017 IEEE Conference on Computer Vision and Pattern Recognition
  (CVPR)}, pages 2261--2269, 2017.

\bibitem{Huang2019CCNetCA}
Zilong Huang, Xinggang Wang, Lichao Huang, C. Huang, Yunchao Wei, Humphrey Shi,
  and Wenyu Liu.
\newblock Ccnet: Criss-cross attention for semantic segmentation.
\newblock {\em 2019 IEEE/CVF International Conference on Computer Vision
  (ICCV)}, pages 603--612, 2019.

\bibitem{Iandola2017SqueezeNetAA}
Forrest~N. Iandola, Matthew~W. Moskewicz, Khalid Ashraf, Song Han, William~J.
  Dally, and Kurt Keutzer.
\newblock Squeezenet: Alexnet-level accuracy with 50x fewer parameters and <1mb
  model size.
\newblock {\em ArXiv}, abs/1602.07360, 2017.

\bibitem{Jia2016DynamicFN}
Xu Jia, Bert~De Brabandere, T. Tuytelaars, and L. Gool.
\newblock Dynamic filter networks.
\newblock In {\em Advances in Neural Information Processing Systems}, 2016.

\bibitem{krizhevsky2009learning}
Alex Krizhevsky and Geoffrey Hinton.
\newblock Learning multiple layers of features from tiny images.
\newblock 2009.

\bibitem{Li2019SelectiveKN}
Xiang Li, Wenhai Wang, Xiaolin Hu, and Jian Yang.
\newblock Selective kernel networks.
\newblock {\em 2019 IEEE/CVF Conference on Computer Vision and Pattern
  Recognition (CVPR)}, pages 510--519, 2019.

\bibitem{lin2017runtime}
Ji Lin, Yongming Rao, Jiwen Lu, and Jie Zhou.
\newblock Runtime neural pruning.
\newblock In {\em Advances in neural information processing systems}, pages
  2181--2191, 2017.

\bibitem{Lin2017ASS}
Zhouhan Lin, Minwei Feng, C.~D. Santos, Mo Yu, B. Xiang, Bowen Zhou, and Yoshua
  Bengio.
\newblock A structured self-attentive sentence embedding.
\newblock {\em ArXiv}, abs/1703.03130, 2017.

\bibitem{liu2018dynamic}
Lanlan Liu and Jia Deng.
\newblock Dynamic deep neural networks: Optimizing accuracy-efficiency
  trade-offs by selective execution.
\newblock In {\em Proceedings of the AAAI Conference on Artificial
  Intelligence}, volume~32, 2018.

\bibitem{Ma2020WeightNetRT}
Ningning Ma, X. Zhang, J. Huang, and J. Sun.
\newblock Weightnet: Revisiting the design space of weight networks.
\newblock In {\em ECCV}, 2020.

\bibitem{Ma2018ShuffleNetVP}
Ningning Ma, Xiangyu Zhang, Hai-Tao Zheng, and Jian Sun.
\newblock Shufflenet v2: Practical guidelines for efficient cnn architecture
  design.
\newblock {\em ArXiv}, abs/1807.11164, 2018.

\bibitem{Park2018BAMBA}
Jongchan Park, S. Woo, Joon-Young Lee, and In-So Kweon.
\newblock Bam: Bottleneck attention module.
\newblock In {\em BMVC}, 2018.

\bibitem{Roy2018ConcurrentSA}
Abhijit~Guha Roy, Nassir Navab, and C. Wachinger.
\newblock Concurrent spatial and channel squeeze excitation in fully
  convolutional networks.
\newblock {\em ArXiv}, abs/1803.02579, 2018.

\bibitem{russakovsky2015imagenet}
Olga Russakovsky, Jia Deng, Hao Su, Jonathan Krause, Sanjeev Satheesh, Sean Ma,
  Zhiheng Huang, Andrej Karpathy, Aditya Khosla, Michael Bernstein, et~al.
\newblock Imagenet large scale visual recognition challenge.
\newblock {\em International journal of computer vision}, 115(3):211--252,
  2015.

\bibitem{Sandler2018MobileNetV2IR}
Mark Sandler, Andrew~G. Howard, Menglong Zhu, Andrey Zhmoginov, and Liang-Chieh
  Chen.
\newblock Mobilenetv2: Inverted residuals and linear bottlenecks.
\newblock {\em 2018 IEEE/CVF Conference on Computer Vision and Pattern
  Recognition}, pages 4510--4520, 2018.

\bibitem{Simonyan2015VeryDC}
Karen Simonyan and Andrew Zisserman.
\newblock Very deep convolutional networks for large-scale image recognition.
\newblock {\em CoRR}, abs/1409.1556, 2015.

\bibitem{Srivastava2014DropoutAS}
Nitish Srivastava, Geoffrey~E. Hinton, A. Krizhevsky, Ilya Sutskever, and R.
  Salakhutdinov.
\newblock Dropout: a simple way to prevent neural networks from overfitting.
\newblock {\em J. Mach. Learn. Res.}, 15:1929--1958, 2014.

\bibitem{Sun2018IGCV3IL}
Ke Sun, Mingjie Li, Dong Liu, and Jingdong Wang.
\newblock Igcv3: Interleaved low-rank group convolutions for efficient deep
  neural networks.
\newblock In {\em BMVC}, 2018.

\bibitem{Szegedy2016RethinkingTI}
Christian Szegedy, Vincent Vanhoucke, Sergey Ioffe, Jon Shlens, and Zbigniew
  Wojna.
\newblock Rethinking the inception architecture for computer vision.
\newblock {\em 2016 IEEE Conference on Computer Vision and Pattern Recognition
  (CVPR)}, pages 2818--2826, 2016.

\bibitem{Tan2018MnasNetPN}
Mingxing Tan, Bo Chen, Ruoming Pang, Vijay Vasudevan, and Quoc~V. Le.
\newblock Mnasnet: Platform-aware neural architecture search for mobile.
\newblock {\em 2019 IEEE/CVF Conference on Computer Vision and Pattern
  Recognition (CVPR)}, pages 2815--2823, 2018.

\bibitem{Vaswani2017AttentionIA}
Ashish Vaswani, Noam Shazeer, Niki Parmar, Jakob Uszkoreit, Llion Jones,
  Aidan~N. Gomez, L. Kaiser, and Illia Polosukhin.
\newblock Attention is all you need.
\newblock {\em ArXiv}, abs/1706.03762, 2017.

\bibitem{Wang2017ResidualAN}
Fei Wang, Mengqing Jiang, Chen Qian, S. Yang, Cheng Li, H. Zhang, Xiaogang
  Wang, and X. Tang.
\newblock Residual attention network for image classification.
\newblock {\em 2017 IEEE Conference on Computer Vision and Pattern Recognition
  (CVPR)}, pages 6450--6458, 2017.

\bibitem{Wang2018NonlocalNN}
X. Wang, Ross~B. Girshick, A. Gupta, and Kaiming He.
\newblock Non-local neural networks.
\newblock {\em 2018 IEEE/CVF Conference on Computer Vision and Pattern
  Recognition}, pages 7794--7803, 2018.

\bibitem{wang2018skipnet}
Xin Wang, Fisher Yu, Zi-Yi Dou, Trevor Darrell, and Joseph~E Gonzalez.
\newblock Skipnet: Learning dynamic routing in convolutional networks.
\newblock In {\em Proceedings of the European Conference on Computer Vision
  (ECCV)}, pages 409--424, 2018.

\bibitem{Woo2018CBAMCB}
S. Woo, Jongchan Park, Joon-Young Lee, and In-So Kweon.
\newblock Cbam: Convolutional block attention module.
\newblock In {\em ECCV}, 2018.

\bibitem{wu2018blockdrop}
Zuxuan Wu, Tushar Nagarajan, Abhishek Kumar, Steven Rennie, Larry~S Davis,
  Kristen Grauman, and Rogerio Feris.
\newblock Blockdrop: Dynamic inference paths in residual networks.
\newblock In {\em Proceedings of the IEEE Conference on Computer Vision and
  Pattern Recognition}, pages 8817--8826, 2018.

\bibitem{Xie2018IGCV2IS}
Guotian Xie, Jingdong Wang, Ting Zhang, Jian-Huang Lai, Richang Hong, and
  Guo-Jun Qi.
\newblock Igcv2: Interleaved structured sparse convolutional neural networks.
\newblock {\em ArXiv}, abs/1804.06202, 2018.

\bibitem{yang2019condconv}
Brandon Yang, Gabriel Bender, Quoc~V Le, and Jiquan Ngiam.
\newblock Condconv: Conditionally parameterized convolutions for efficient
  inference.
\newblock In {\em Advances in Neural Information Processing Systems}, pages
  1307--1318, 2019.

\bibitem{yu2018slimmable}
Jiahui Yu, Linjie Yang, Ning Xu, Jianchao Yang, and Thomas Huang.
\newblock Slimmable neural networks.
\newblock {\em arXiv preprint arXiv:1812.08928}, 2018.

\bibitem{Yuan2018OCNetOC}
Y. Yuan and Jingdong Wang.
\newblock Ocnet: Object context network for scene parsing.
\newblock {\em ArXiv}, abs/1809.00916, 2018.

\bibitem{Zhang2020ResNeStSN}
Hang Zhang, Chongruo Wu, Zhongyue Zhang, Yi Zhu, Zhi-Li Zhang, Haibin Lin, Yu e
  Sun, Tong He, Jonas Mueller, R. Manmatha, M. Li, and Alex Smola.
\newblock Resnest: Split-attention networks.
\newblock {\em ArXiv}, abs/2004.08955, 2020.

\bibitem{Zhang2017InterleavedGC}
Ting Zhang, Guo-Jun Qi, Bin Xiao, and Jingdong Wang.
\newblock Interleaved group convolutions.
\newblock {\em 2017 IEEE International Conference on Computer Vision (ICCV)},
  pages 4383--4392, 2017.

\bibitem{Zhang2018ShuffleNetAE}
Xiangyu Zhang, Xinyu Zhou, Mengxiao Lin, and Jian Sun.
\newblock Shufflenet: An extremely efficient convolutional neural network for
  mobile devices.
\newblock {\em 2018 IEEE/CVF Conference on Computer Vision and Pattern
  Recognition}, pages 6848--6856, 2018.

\bibitem{Zhao2018PSANetPS}
Hengshuang Zhao, Yi Zhang, Shu Liu, J. Shi, Chen~Change Loy, D. Lin, and J.
  Jia.
\newblock Psanet: Point-wise spatial attention network for scene parsing.
\newblock In {\em ECCV}, 2018.

\bibitem{Zoph2018LearningTA}
Barret Zoph, V. Vasudevan, Jonathon Shlens, and Quoc~V. Le.
\newblock Learning transferable architectures for scalable image recognition.
\newblock {\em 2018 IEEE/CVF Conference on Computer Vision and Pattern
  Recognition}, pages 8697--8710, 2018.

\end{thebibliography}
}

\end{document}